\documentclass{article}

 \usepackage[final]{neurips_2020}

\usepackage[utf8]{inputenc} 
\usepackage[T1]{fontenc}    
\usepackage{hyperref}       
\usepackage{url}            
\usepackage{booktabs}       
\usepackage{amsfonts}       
\usepackage{nicefrac}       
\usepackage{microtype}      
\usepackage{graphics} 
\usepackage{graphicx, color}

\title{Hidden Technical Debts for Fair Machine Learning in Financial Services}

\author{%
    Chong Huang\thanks{Equal contribution} \\
  FICO AI Research\\
  San Jose, CA 95110 \\
  \texttt{chonghuang@fico.com} \\
  \And
    Arash Nourian\footnotemark[1] \\
  FICO AI Research\\
  San Rafael, CA 94903 \\
  \texttt{arashnourian@fico.com} \\
  \And
    Kevin Griest \\
  FICO AI Research\\
  San Rafael, CA 94903 \\
  \texttt{kevingriest@fico.com} \\
}

\begin{document}

\maketitle

\begin{abstract}
The recent advancements in machine learning (ML) have demonstrated the potential for providing a powerful solution to build complex prediction systems in a short time. However, in highly regulated industries, such as the financial technology (Fintech), people have raised concerns about the risk of ML systems discriminating against specific protected groups or individuals. To address these concerns, researchers have introduced various mathematical fairness metrics and bias mitigation algorithms. This paper discusses hidden technical debts and challenges of building fair ML systems in a production environment for Fintech. We explore various stages that require attention for fairness in the ML system development and deployment life cycle. To identify hidden technical debts that exist in building fair ML system for Fintech, we focus on key pipeline stages including data preparation, model development, system monitoring and integration in production. Our analysis shows that enforcing fairness for production-ready ML systems in Fintech requires specific engineering commitments at different stages of ML system life cycle. We also propose several initial starting points to mitigate these technical debts for deploying fair ML systems in production. 
\end{abstract}

\section{Introduction}

Machine learning (ML) systems play an increasingly important role in making various decisions in financial industry ranging from customer on-boarding for a loan to detecting fraudulent transactions. These systems utilize a massive amount of historical data to build complex prediction systems that provide valuable insights for decision making, thus improve efficiency and reduce costs. Despite many advantages that ML systems can potentially provide, the financial technology (Fintech) industry is one of the few industries that has not fully utilized the latest advancements in this space. One major challenge for adopting ML systems in Fintech is that it is highly regulated and often requires careful scrutiny of models used to make high stake decisions. The lack of explainability in most black-box ML systems brings the potential risk that the ML based systems may identify patterns that exist because of inadvertent discrimination and/or exclusionary behavior. Indeed, these unintended but harmful discriminatory behaviors have been observed in many such systems \citep{vigdor2019apple, bowyer2020criminality, dastin2018amazon}. 
Recent advancements in explainable AI and more transparent/self-explaining ML models have put ML based systems on the forefront of transforming financial industry \citep{hagras2020banking}. These new advancements have also highlighted challenges around how to evaluate fairness in the ML systems and mitigate potential unfair treatments specially in a fast changing production setting – a real concern for Fintech companies considering whether to put advanced ML systems into production.

\begin{figure}[ht]
\centering
\includegraphics[width=0.9\columnwidth]{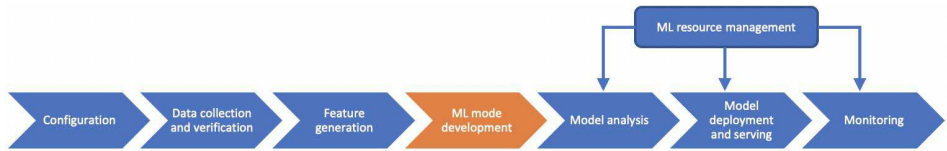}
\caption{Simplified example of ML system in production}
\label{fig:pipeline}
\end{figure}

The notion of technical debt is introduced in \citep{cunningham1992wycash} as a metaphor to describe the costly long-term issues emerged by choosing an expedient solution in the short-term and has been used to capture potential risk factors in ML system design \citep{sculley2015hidden}. In this paper, we argue that there exists a variety of hidden technical debts and challenges for ML practitioners to consider when building a production ready fair ML system for Fintech. These systems not only have all of the traditional issues with respect to mitigating fairness challenges in ML models but also suffer from Fintech-specific fairness issues specially in production environment that are often ignored. These debts may be difficult to detect since they exist at the system level across the whole life cycle of ML system when pushed in production. Hidden technical debt is a real concern for ML systems in production because it accumulates silently and can become extremely expensive to payoff.  Our analysis reveals that typical methods for addressing fairness in ML models may not be sufficient to solve the challenges at the system level because ML model is just a small component of the system in production (Figure \ref{fig:pipeline}).

ML systems' discriminatory behavior may be due to a variety of reasons. For example, some ML systems may give privileges to certain subgroups of the data due to the characteristics associated with the dataset used for training the ML system (e.g., too few data samples for a certain subgroup, or biases that are embedded in the data because of historical policies and decisions). Other ML systems might have fairness issues due to biases inherited from the learning algorithm, human intervention during the model development, and interactions with other components in the deployment process. As a result, ensuring fairness end-to-end in ML systems is challenging in many aspects. Discriminatory behavior can creep in not only from ML models but also from other components in the system such as data collection, data augmentation, post-model deployment and monitoring etc (Figure \ref{fig:pipeline}). All of these components can contribute to accumulating technical debts that make building fair ML systems in production costly. Although some of the technical debts can be paid off by adopting what we introduce as \textit{fairness by design} which puts fairness at the core of building ML systems when refactoring code, designing components for a decision flow, and providing fairness specific tests for production environments. These techniques may not be applicable to other debts at the ML system level. Therefore, designing a fair ML system without careful consideration of the hidden technical debts upfront will bring issues that accumulate and become costly to address in the long run.

\subsection{Contributions}
The scope of this paper is not focused on offering novel solutions to the fair ML problem. Instead, we seek to highlight the hidden technical debts for developing and deploying fair ML systems with \textit{fairness by design} paradigm in production and increase the community’s attention to the difficult challenges that should be considered in large scale production environment over the long-term. The main contributions of this paper are: 1) identifying hidden technical debts in building fair ML system at different stages of the life cycle of ML  in production; 2) highlighting challenges for enforcing fair ML systems in production environment; and 3) proposing \textit{fairness by design} with starting points to address the challenges for enforcing fairness in ML systems used in Fintech industry. 
\section{Data Related Technical Debts}
As ML systems rely on data to learn, the input data can influence their behavior in many ways. A detailed survey of potential biases in data can be found in \citep{mehrabi2019survey}. In this section we mainly focus on technical challenges and debts related to the data used to build fair ML systems in Fintech industry.   

\subsection{Mitigate sampling bias}
Due to social and historical biases in the related use cases, it is possible that there exists sampling biases in the datasets collected to train the ML systems. The sampling bias may impact the ML system in an unpredictable manner which can result in biased outcome. For example, \citep{harney2018large} shows that loan applications of black applicants are denied at a higher rate in history. If historical loan application data is used to train the ML system for risk analysis, the information about whether or not black applicants would have defaulted had their application been approved may be heavily underrepresented in the dataset since less black applicants are approved compared to the other races. Using a dataset which has potential sampling bias to train the ML system could lead to biased or even sub-optimal decision-making strategies. Therefore, sampling bias mitigation methods should be designed carefully to avoid adding technical debt to the system which makes building fair ML systems more costly. Addressing sampling biases become even more critical when multiple datasets are used or merged to train the ML system. In the case of a merged dataset, it is important that the information associated with all the original datasets used in merging are passed as a metadata in the merged dataset for further consideration of potential sampling bias.

An ideal but naive approach to mitigating sampling bias is to actively accept more of the underrepresented population until we have a balanced dataset for development. However, this is both time-intensive and brings a significant financial burden to the lender so is infeasible in production. Upsampling can help balance the number of samples but the diversity of the data is limited. Reject inference \citep{banasik2007reject} is a common technique to address this issue. 
Traditional reject inference requires building models on the dataset to augment the training data by inferring the performance of rejected samples. While this approach reduces the under-represented bias by modeling the ``through-the-door'' population, it can also magnify other biases that are encoded in the known population's label. 
For example, if the accepted black applicants are more likely to default due to biased decisions such as predatory loan pricing, an inference model without fairness constraint may generate estimations that are even less favorable to the black applicant who has been rejected. Therefore, the inference process should be designed carefully to eliminate extra biases introduced into the ML system.  

\subsection{Isolate proxy variable}
Given the variety of features used for building ML system, it is challenging to establish that a biased decision outcome is solely driven by “non-protected” features rather than protected features. In \citep{ross2008mortgage}, the authors found that minority borrowers were more often encouraged to apply FHA loans, which are considered to be more expensive to finance. In this scenario, the loan type information can be considered as a proxy for race and can carry biased information if not treated properly. These proxy variables are quite common in real datasets. For example, in the widely used German Credit dataset \citep{Dua:2019}, both ``Risk'' and ``Sex'' are positively correlated with the ``Housing'' feature. If the correlation is strong, ``Housing'' becomes a proxy variable for the protected ``Sex'' feature.
Other studies have raised concerns that income level \citep{weinberg2007earnings} and zip code \citep{datta2017proxy} can be proxy features for gender and race. Unfortunately, it is very difficult (in some cases almost impossible) to separate the proxy variables' influence on the target variable with their correlation with protected features. Including these proxy variables in the dataset to train ML systems may result in biased decisions, even if the protected features are excluded explicitly. Therefore it is crucial to investigate the dataset for potential proxy variables as well as leaking such variables through feedback loops in a possible retraining process. Including proxy variables without careful studying of their influence on ML system fairness is a technical debt, which may be difficult to identify for the ML system in production specially when the ML system is run as a micro-service interacting with other services or applications.

\begin{figure}[ht]
\centering
\includegraphics[width=0.37\columnwidth]{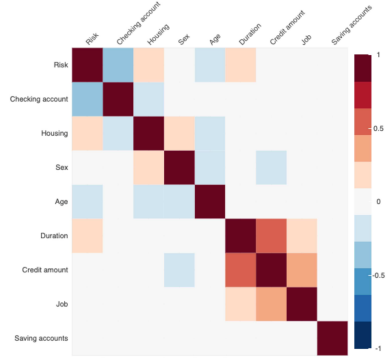}
\caption{Pearson correlation between features for German Credit dataset}
\label{fig:correlation_german}
\end{figure}


\subsection{Handle missing data}
In many real-world application settings, missing data is a common challenge to tackle for data mining and ML model development \citep{lin2020missing}. In Fintech industry, missing data can be present in a variety of applications such as credit scoring and fraud detection. 
In some cases, discarding data with missing value seems to be a simple solution but may lead to sampling bias that could violate fairness constraints of the ML system. In many other cases, missing value imputation \citep{lin2020missing} is used to estimate missing data. However, many imputation methods are trained based on the dataset and may introduce extra unpredictable biases to the dataset. For example, using mean value imputation could potentially shift the distribution of the minority group closer to the majority group of the dataset. Furthermore, after the model has been deployed in production, evaluating fairness metrics on data samples with missing protected feature information (e.g., gender or sex) could also be challenging for many group fairness metrics (e.g., demographic parity and equalized odds \citep{hardt2016equality}) which require access to the protected feature label. 
\section{Fairness Metrics Related Technical Debts}

Another area where technical debt can accumulate is the configuration of metrics for evaluating fairness of ML systems. To investigate whether there exists discrimination in the output of ML models, numerous fairness metrics have been introduced to capture fairness via mathematical formulation, such as equalized odds and equal opportunity \citep{hardt2016equality}, positive predictive value \citep{chouldechova2017fair}, counterfactual fairness \citep{kusner2017counterfactual}, earthmover distance \citep{dwork2012fairness}, and generalized entropy index fairness \citep{speicher2018unified}. On one hand, these metrics provide a set of measurements that investigates fairness from different perspectives. On the other hand, the variety of fairness metrics has introduced new challenges in determining which metric is the most relevant for different use cases in Fintech industry. 

\subsection{Lack of consensus among metrics}
Interpreting the impact of different fairness metrics is a difficult task given that many of them are not mathematically compatible for the fairness definition \citep{corbett2018measure}. In the heavily regulated Fintech industry, it would be challenging to pick the right fairness metrics for the task at hand and justify the preference of one metric over the others specially when the production environment could change over time. Furthermore, due to the various requirements from regulatory bodies, the ML systems are subject to constant auditing. For regulators and auditors, it is important to understand the fairness metrics used both in evaluating models and in the production as well as the relationship between measurements and the compliance with fair decision regulations. In such scenarios, intuitive and effective metrics are preferred over complex mathematical formulation. Due to the complexity of the integrated ML systems in a production setting and the backward compatibility requirements, changing fairness metrics in production is expensive. Selecting the most suitable metrics and being able to justify the metrics and the outcome of the ML system to regulators is another important step to avoid accumulating technical debt for building fair ML system. It is also crucial to provide evidence to show that for a specific use case, no other fairness metric performs better at capturing discrimination of the ML system without harming the business use case. Furthermore, given the fact that there exists significant disagreement in people’s perception of “unfair” treatment, coming up with a consensus metric to capture the heterogeneity in people’s understanding of discrimination in various use cases would also be a challenge in enforcing fair ML in production. Therefore, having consensus on justification of fairness metrics for the specific tasks by utilizing domain expertise is an important hidden debt that needs to be considered upfront when designing ML systems. 

\subsection{Fairness-performance trade-off}
Strictly enforcing a fairness constraint, which only focuses on the risk of discrimination for the protected group without considering potential benefits of the ML systems, may also hurt the overall utility of the ML system. For example, enforcing complete demographic parity fairness constraint on a ML model may often cripple the utility of the model \cite{hardt2016equality}. Given the dynamic nature of environment change in ML systems pushed in production, it is very difficult or almost impossible for some cases to completely eliminate biases while guaranteeing good utility for a decision maker. However, it is possible to engineer a production friendly ML system that is reasonably fair and performs well on other objectives for decision making. It is important to document and acknowledge the potential edge cases such engineered systems can not handle for auditing purposes.

In reality, a decision maker might have several competing objectives to consider. Therefore, rather than enforcing fairness as a strict ``hard'' constraint (e.g., enforcing demographic parity), it would be beneficial for the decision maker to view fairness as a ``soft'' constraint (e.g., setting up a tolerable threshold for the fairness metrics) and thus shifting the focus from completely removing bias based on quantified fairness metrics to designing a ML system that offers the best trade-off between competing objectives. With such blueprints for measuring fairness in production grade ML systems, setting up a threshold for specific fairness metrics to determine whether the bias in the ML system meets the requirements could be a potential widely adopted solution. 
However, determining the right metrics and thresholds to answer the question of ``how fair is fair enough?'' for different use cases and contexts will be a difficult task and the choices will have strong influences on the system behavior in the long term. The accumulated influence of fairness metrics can create technical debts that are increasingly difficult to address later in the production environment.

\subsection{Identifying counterfactual}
With the potential biases inherited from data collection, historical discrimination and human intervention (e.g., manual overrides of decision), it is not sufficient to evaluate fairness only based on statistical metrics, such as parity in selection or error rates. It would be beneficial to have a counterfactual setting to drill down to the root cause of discrimination and disentangle the correlation between proxy variables for both target and protected features by modeling the causes and effects via directed acyclic graph \cite{kusner2017counterfactual}. Given the complexity of discovering causal relationship for modern ML models, it would be challenging to implement this in real-world ML decision systems. Such relationships could constantly change the causes in an environment with many potential sudden shifts.

\subsection{Metrics precision vs. cost}
For an ML system in production, it is often costly to evaluate various fairness metrics on large datasets due to computation complexity of the metrics and the amount of data samples needed for evaluating such metrics. Due to limited resources in practice, the ML practitioners may have to make a trade-off between precise fairness evaluation (typically costly evaluation processes) versus alternative approximations. These approximations are often not carefully evaluated/tested and could potentially cause serious issues that could go undetected for a long time. Therefore, simple and yet effective metrics and evaluation process is of paramount importance for fairness evaluation on large datasets in production environment. Otherwise, less precise metrics may be used in production due to cost constraints and they may not be sufficient to discover discriminatory behaviors completely. Furthermore, in a production environment, costs associated with fairness evaluation on inference is also a concern if the metrics used need to be evaluated on large sample size. Thus, balancing the trade-off between precision of fairness evaluation and the cost of evaluation can help reducing hidden technical debt for building fair ML systems in production. 
\section{Production Monitoring Related Technical Debts}
Deployed ML systems for decision making often interact directly with the external world or other applications (e.g., taking a stream of loan applicants' data and providing the predicted probability of default). With a constant changing environment where applications are spawned by various up-stream/down-stream services or different users, this creates several challenges and potential technical debts for monitoring fairness guarantees of the ML system for decision making in production.  

\subsection{Issues related to dynamic nature of production environment}
Decision making often requires a decision threshold to predict the applicant will default or not, to detect whether the voice is from a fraudster, and to offer or not offer a credit line increase. The canonical approach is to pick a threshold based on system performance on certain metrics, such as accuracy or precision. However, the thresholds are often set manually to optimize the metrics. If a model updates on new data or the distribution of incoming data in production is different from the data used for model development, the manually set threshold may not enforce the desired fairness constraint for the new input. Manually checking and updating the thresholds in the production may be time consuming and prone to errors. One possible mitigation strategy for this problem is to use ML techniques such as Reinforcement Learning to adjust thresholds that achieve a trade-off between fairness and performance metrics in a dynamic environment. Thus, building the fair ML system without taking the dynamic environment of the ML system into consideration is another technical debt that needs special attention. 

\subsection{Monitoring fairness metrics in production}
Constant and comprehensive monitoring of ML systems in real-time production environment is critical for long-term system reliability. One important question for monitoring fair ML systems is the degree of the monitoring in production. Unlike ML model development process where test datasets are often used, test datasets strictly used for fairness evaluations are not always readily available given that many ML systems in production evolve over time. Furthermore, the changes in the input data bring extra challenges (e.g., evolving consumer behavior may affect ML systems fairness). How to monitor fairness for dynamically evolving ML systems in production remains a challenging task but needs to be baked into fair ML system design to avoid technical debts. We propose the following aspects as starting points and the key pieces of \textit{fairness by design} for ML systems:
\begin{itemize}
    \item Monitoring ML system output bias: 
    the first level of monitoring involves a set of fairness metrics tests that each model needs to pass based on a holdout test data and random simulated data. This is by no means a comprehensive test but it is a useful initial evaluation process to detect unfair systems. Also, changes in fairness metrics may indicate issues that require attention. Investigating the disparity of various performances and fairness metrics can also help detect discriminatory behavior of the system. 
    \item Limiting certain decisions/actions: for an ML system that makes decisions in a live production environment, such as loan origination, it can be useful to set and enforce limits on the decisions made on the protected group as a sanity check. For example, one could limit the rejection rate for the protected group. If the system hits a limit for the rejection rate threshold set by a domain expert, automated alert is triggered and the responsible parties receive a notification for presence of potential  bias in the system that needs further investigation. 
    
    \item Monitoring changes for input data and data processors:
    data is often pre-processed before it is fed to a ML model, it is also useful to monitor the input features for biases, to make sure that the distribution of values is relatively stable. It would be beneficial to raise alert if the input features, various transformations, or data augmentation were to vary significantly, as this would cause unwanted changed behavior for the ML system. Also, the data processors used at every step of a ML pipline should be thoroughly monitored, tested and audited to check for potential biases introduced for the down-stream ML systems. 
\end{itemize}
\section{Debiasing Related Technical Debts}
To ensure the output of a ML model is fair, a variety of algorithms have been proposed to mitigate biases in ML systems based on mathematical fairness metrics. These algorithms enforce fairness constraints at different stages of ML model development including pre-processing \citep{zemel2013learning, madras2018learning, lahoti2019ifair, feldman2015certifying}, post-processing \citep{hardt2016equality}, and learning fair model \citep{zhang2018mitigating, agarwal2018reductions}. The diversity of bias mitigation algorithms provides a wide range of choices for users to mitigate biases in ML systems but also brings several hidden technical debts for building fair ML system.

\subsection{Consensus and guideline on bias mitigation algorithms}
Given the large pool of bias mitigation algorithms, it is difficult to determine which algorithm works best for different use cases in production and justify why one mitigation algorithm is preferred over another without establishing coherent guidelines to use in the production environment. Furthermore, for compliance and regulation purposes, explaining the deployed debiasing techniques to regulators is another important technical debt to consider when choosing bias mitigation algorithms. As ML system in production is very costly to modify, it is also crucial to ensure that the bias mitigation algorithm is compliance friendly and generates required real-time auditory reports without harming the business use. Introducing a general guideline for the pros and cons of each bias mitigation algorithm and the applicable use cases would be a starting point for tackling this issue. 

\subsection{Configuration of bias mitigation algorithms}
Another area that may accumulate technical debts is the configuration of the bias mitigation methods. Each bias mitigation algorithm may have different configuration options, including which features to use, how data is processed, and various algorithm specific configuration parameters. It is challenging to optimize, reason, verify and test the set of configuration variables in a live system. Moreover, in a production environment, the combination of configurations of other components and the bias mitigation modules can be very large and may have accumulated effects on the outcomes while incurring huge computational cost if not engineered properly. The error prone nature of manual configuration can be costly and hard to identify, leading to a significant cost in production environment. 

\subsection{Accounting for adversarial scenarios}
Due to the potential disparity in treating different subgroups, malicious users may take advantage of such disparity in the system to increase the likelihood of obtaining their favorable outcome \citep{miller2018gamming}. Thus, the correction of biased behavior, either through pre-processing, post-porcessing or learning fair models, should not incentivize malicious users to game the system. For instance, if the debiasing algorithm has a higher probability of offering favorable outcome to a protected group, the process for debiasing should also prevent the users from extracting potential vulnerabilities to their benefits. Such debiasing scanning can result in adversarial actions such as users hiding their actual protected feature (e.g., sex) in exchange for a favorable outcome. The cost of successfully gaming the system is extremely expensive and may lead to significant issues in production environment. 

\subsection{Consistent debaising behavior}
For ML systems in production, consistency and stability of the system are important requirements. Thus, for any debiasing algorithm deployed in production, the output of the ML system after debiasing is expected to perform consistently over time. Given the randomness and unstable behavior of some of the bias mitigation algorithms \citep{lee2020algorithmic} and the close interaction with other components, it might be challenging to obtain strictly consistent debiased outcome in a dynamic production environment. This may create various problems for regulatory parties to audit the system and justify the proper use of these bias mitigation algorithms. 
\section{CI/CD Related Technical Debts}
In a production environment, one ML system (i.e., an entire pipeline) could be a small part of a large ecosystem of pipelines that collaborates together in delivering a precision decision. The ML system has to be integrated tightly with other components in the production environment. For example, ML systems may be connected with up-stream data provider which generates real-time data streams for prediction and a down-stream rule based decision execution engine for making the decisions. The close interactions between different subsystems in the production environment makes the continuous integration and deployment of fair ML systems an area that hidden technical debt can accumulate. 

\subsection{Integration of multiple fairness evaluation methods in production}
Since ML systems in production are not living organisms in isolation and are typically chained with other components, these systems can be affected by the behavior of other interacting components. Integrating different components in the production environment requires supporting code and attached schema to get data in and out of each component. It is costly in the long run to simply ``glue'' different components together \citep{sculley2015hidden}. Furthermore, in the production environment, the ``glued'' components make identifying causes of biased behavior difficult. Running unit tests works for individual components but fails to capture the aggregated effects in a heavily integrated system. Thus, investigating the influence of other components' on the fairness properties of ML systems for fairness evaluation is a challenging but required task. Creating a standard procedure to assess the influence of the changes in other components on fairness in production as well as designing integration tests that proactively embed fairness can help reduce technical debt in the long run.

\subsection{Integration of bias mitigation in production}
For a ML system in production environment, integrating the bias mitigation algorithms should also take the influence of other integrated components into consideration. Once the discriminatory behavior has been identified, it is important to single out the components that contribute to the biased outcomes and correct their behavior. Multiple reconfiguration, retraining or rebuilding a model in production is costly, challenging, and may change the behavior of other components. Modifying the system due to fairness concerns in an integrated production environment should be carefully investigated to avoid causing problems for up-stream and down-stream applications.

In a production environment, the bias mitigation module along with other components in the production environment should be stored in a repository from which the developed fair ML systems could be restored quickly and seamlessly in production. Managing, trouble shooting, and recovering these systems are difficult and costly. Moreover, these accumulated versions of different components can create more debt due to increasing difficulties to maintain backward compatibility. Also, testing the system integration often requires extensive unit tests and expensive end-to-end integration tests. 
\section{Reducing Hidden Technical Debt via Fairness by Design}
To reduce potential technical debts for fair ML systems, we propose using \textit{fairness by design} paradigm for building production ML systems. At its core, fairness by design means integrating fairness evaluation and bias mitigation into the system design, development and deployment life cycle. It shouldn't be a simple isolated model fairness analysis or add-on to the existing procedure. We resort to the following strategies as starting points:

\begin{itemize}
    \item Build ML systems with fairness baked into the design, development, and production deployment.
    \item Carefully monitor and logging data input, such as potential proxy variable as well as the corresponding meta data. 
    \item Consolidate the metrics, evaluation process, as well as the bias mitigation algorithm that are appropriate for different use cases and regulations.  
    \item Constantly monitor and alert potential biased behavior of the ML system in a dynamic production environment. 
    \item Create standard procedures to assess the influence of changes in other components on fairness in production and design unit/integration tests that proactively embed fairness.
\end{itemize}

\section{Conclusion}
In this paper, we highlighted some of the technical debts for building and deploying fair ML systems in production from the perspective of data, fairness metrics, debiasing methods, integrating and monitoring such systems in production. Guaranteeing end-to-end preservation of a fair metric is not a simple task and requires specific commitments in different stages of ML system life cycle. Our analysis shows a need for \textit{fairness by design} approach when developing and deploying ML with fairness properties in production. 

We hope that this paper may serve as an initial endeavor to encourage broader discussions between academia and industry with respect to developing and deploying fair ML systems for decision making in Fintech. Developing theoretical frameworks considering real-world problems has several challenges such as continuous integration, monitoring and testing discrimination of ML systems in production. More comprehensive guidelines for choosing fairness metrics and bias mitigation algorithms are essential for successful deployment of fair ML systems. Properly treating these technical debts and challenges can assist Fintech industry to utilize fair ML systems more broadly in real-world problems.







\bibliographystyle{plainnat} 

\bibliography{reference}

\end{document}